\documentclass{article} 
\usepackage{iclr2027_conference,times}

\usepackage{amsmath,amsfonts,bm}

\def\eqref#1{equation~\ref{#1}}

\def\1{\bm{1}}

\DeclareMathAlphabet{\mathsfit}{\encodingdefault}{\sfdefault}{m}{sl}
\SetMathAlphabet{\mathsfit}{bold}{\encodingdefault}{\sfdefault}{bx}{n}

\usepackage{hyperref}
\usepackage{fontawesome5}
\usepackage{url}

\usepackage{booktabs}
\usepackage{graphicx}
\usepackage{algorithm}
\usepackage{algpseudocode}
\usepackage{xcolor}
\definecolor{mydarkblue}{rgb}{0,0.08,0.45}
\hypersetup{colorlinks=true, linkcolor=mydarkblue, citecolor=mydarkblue, urlcolor=mydarkblue}

\usepackage{rotating}
\usepackage[most]{tcolorbox}
\usepackage{listings}
\newtcolorbox[auto counter]{mybox}[2][]{
  enhanced, breakable,
  skin first=enhanced, skin middle=enhanced, skin last=enhanced,
  coltitle=white, fonttitle=\bfseries,
  title=Box \thetcbcounter: #2, #1
}

\algrenewcommand{\algorithmiccomment}[1]{\hfill$\triangleright$ #1}
\title{Code to Control: Synthesizing Parameterized Reactive Controllers}

\author{%
Zergham Ahmed$^{1}$ \quad Joshua B. Tenenbaum$^{2}$ \quad Chris Bates$^{1,3}$ \quad Samuel J. Gershman$^{1}$ \\[2pt]
\normalfont $^{1}$Harvard University \\
\normalfont $^{2}$Massachusetts Institute of Technology \\
\normalfont $^{3}$Florida Institute for Human and Machine Cognition \\[6pt]
\normalfont\small \faEnvelope\,: \texttt{zerghamahmed@g.harvard.edu} \\
\normalfont\small \faGithub\,: \url{https://github.com/ZerghamAhmed/code-to-control} \\
\normalfont\small \faGlobe\,: \url{https://zerghamahmed.github.io/code-to-control}
}

\iclrfinalcopy 
\begin{document}

\maketitle
\lhead{Preprint.}

\begin{abstract}
Recent LLM-based approaches to control either invoke a language model to
select actions or synthesize world models that require planning at every
decision, introducing latency that can limit real-time use. We introduce
Code to Control, an approach that synthesizes
Python controllers which execute directly as policies. Code to Control
separates program structure from parameters. An LLM synthesizes
the controller structure, while derivative-free search fits its parameters
for continuous control using feedback from the environment. Once learned, the resulting controllers require neither LLM inference nor planning at decision time, enabling real-time gameplay and, under our timing protocol, faster action selection than a PPO policy. Across a suite of Atari games, Flappy Bird, and MuJoCo tasks, Code to Control outperforms planning-based program synthesis methods, remains competitive with deep reinforcement learning while using
fewer environment interactions, transfers across substantial
changes in environment dynamics, and scales to complex locomotion tasks. 
\end{abstract}

\section{Introduction}

A central goal in building general-purpose agents is to develop systems that
can act in real time, learn quickly, and generalize flexibly. Language models
used directly as policies provide a flexible mechanism for action selection
\citep{yao2023react}, but require language-model inference at every decision.
Theory-based reinforcement learning (TBRL) and related programmatic world-model
approaches are sample efficient and flexible
\citep{tsividis2026human,ahmed2025synthesizing,ahmed2026learning,
tang2024worldcoder,dainese2024generating,piriyakulkij2026poe}, but lack fast
execution because selecting an action requires planning with the learned
model. Correct world models can also be rather complex for some domains, and small synthesis errors can be challenging to debug and easily compound during rollouts.  

\begin{figure}[t]
\centering
\includegraphics[width=\textwidth]{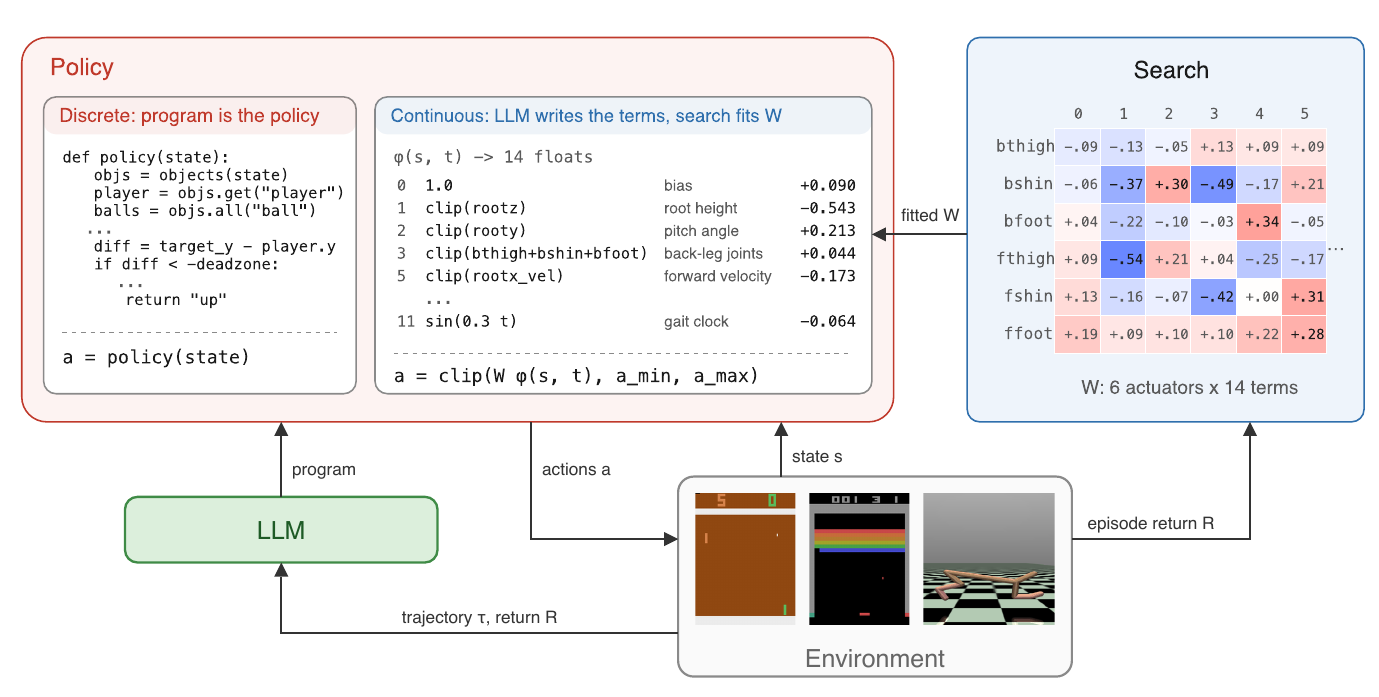}
\caption{\textbf{Code to Control} schema. The LLM synthesizes the
programmatic structure of the controller, while numerical parameters can be
fit separately from environment feedback when useful, especially in
continuous-action domains. The synthesized structure may be a complete
executable controller or a parameterized representation whose numerical
parameters are optimized separately. In our continuous-control experiments,
we instantiate this representation as a feature map $\phi(o,t)$ with a fitted
matrix $W$ mapping features to actuator commands.}
\label{fig:concept}
\end{figure}

To address these challenges, we develop a structured programmatic policy
synthesis and parameter fitting framework, which we call Code to Control. The framework
separates the synthesized program structure from numerical parameters that are fit directly from environment returns. The LLM synthesizes
the controller structure, while derivative-free search fits these numerical
parameters. For example, in motor-control domains, the LLM synthesizes a
feature map whose terms compute interpretable quantities of the observation
and timestep, and derivative-free random search fits the matrix mapping those
features to actuator torques \citep{mania2018simple}. The nature of the free
parameters varies somewhat by domain, and our current instantiation provides
high-level guidance on how to construct them.

The LLM is given neither the environment dynamics nor a demonstration. Unlike
prior work \citep{liang2023code}, Code to Control learns from interaction and
reward rather than relying on a task-specific natural-language command and
predefined control primitives. After synthesizing an initial controller, our
agent plays a game by calling the executable program on the observation, and
the program outputs the action. The LLM then revises the program from its
gameplay trajectory and the reward it produced. Our method is especially
well-suited to domains where relatively simple, reactive-style policies can be
highly effective and where constructing complete world models for planning
would be laborious and offer minimal performance benefits.

\textbf{Our contributions are as follows.}
First, we introduce Code to Control, a planning-free framework for
interpretable programmatic policy synthesis that separates synthesized
controller structure from numerical parameters that can be fit from
environment feedback. Second, we conduct a thorough empirical evaluation across
six Atari games, Flappy Bird, and ten MuJoCo \citep{todorov2012mujoco} tasks,
comparing against planning-based program synthesis, an LLM acting directly as
the policy, and deep reinforcement learning. Code to Control outperforms
planning-based program synthesis methods on all six Atari games and on nine of
ten MuJoCo tasks. It also exceeds PPO trained for 20 million steps on three of
six Atari games while using fewer than 60k environment steps on most games.
On six MuJoCo tasks, the controllers also score above an oracle planner given
the true simulator, suggesting that planning itself, rather than model error
alone, can be a bottleneck. On Flappy Bird, we provide a transfer and adaptation
study across changes in environment dynamics, showing zero-shot transfer on
several perturbations and additional adaptation through parameter refitting
while keeping the synthesized program fixed. Third, we provide an analysis of
the computational cost of action selection, measuring program execution,
planning latency, and LLM API latency. Our method is
faster by construction than planning and LLM-based baselines, and even
executes faster than PPO.

\section{Method}
\label{headings}

\subsection{Problem Formulation}

We take each domain to be a finite-horizon Markov decision process (MDP) with observation
space $\mathcal{O}$, action space $\mathcal{A}$, transition function $T$, reward function $R$, and
horizon $H$. At timestep $t$ the agent observes $o_t \in \mathcal{O}$, chooses an action
$a_t \in \mathcal{A}$, and receives a scalar reward $r_t$. Episodes last at most $H$ timesteps and
may stop earlier if an event terminates the environment. Thus, a finished episode leaves a
trajectory $\tau = \big( (o_t, a_t, r_t, o_{t+1}) \big)_{t < T_\tau}$, a sequence of
$T_\tau \leq H$ transitions, whose return $r(\tau) = \sum_{t < T_\tau} r_t$ is the undiscounted sum
of the rewards collected along it. This trajectory of transitions is shown to the LLM to use to
synthesize the policy. The optimization objective is $\max_{\hat{\pi}} \; \mathbb{E}_{\tau \sim \hat{\pi}} \big[\, r(\tau) \,\big]$.


\paragraph{Policy Synthesis using LLM.} We sample a large language model's space of policies expressible as source code, $\Pi$. A
candidate is a program $\hat{\pi} : \mathcal{O} \times \mathbb{N} \to \mathcal{A}$ carrying the
current observation and the timestep to an action, which in discrete domains reduces to
$\hat{\pi} : \mathcal{O} \to \mathcal{A}$. Here, $\Pi$ is programs in Python, a general-purpose language. The LLM supplies both the prior over
$\Pi$ and the proposals drawn from it. Section~\ref{sec:policyrep} gives the representation used in
each setting.

\subsection{Code to Control}
\label{sec:policyrep}

\paragraph{Overview.} Our agent starts with a text-encoded representation of the observation. For discrete domains, it synthesizes a policy function, also referred to as controller, which takes the observation as input and returns an action. The action is executed in the environment and the environment returns the updated observation. The policy function is revised by our agent while it interacts with the environment. The method of revision differs based on the class of domains. For continuous domains, it
instead writes the terms of a feature map $\phi(o,t)$, and search fits the matrix W that maps those terms to a continuous action output space. During interaction with the environment, the weights are revised by a search. 

\paragraph{Discrete domains.} The observation includes the set of objects, with their object
type as the keys, the centre and the size of each instance's bounding box, and a velocity computed from consecutive frames, which can be read as an object-oriented MDP \citep{diuk2008object}. For
example, if the type is \texttt{ball}, then \texttt{ball} holds a list of centers
$[[x, y], \dots]$, one per instance, \texttt{ball\_size} holds the matching
$[[\textit{width}, \textit{height}], \dots]$, and \texttt{ball\_velocity} holds the per-tick
deltas $[[dx, dy], \dots]$.

\paragraph{Continuous domains.} The observation input to our system includes joint positions
together with joint velocities, which is what determines the dynamics. For example, on HalfCheetah the key \texttt{bthigh} holds the back-thigh joint angle and \texttt{bthigh\_angular\_velocity}
holds its rate.

In discrete domains the policy is a program $\hat{\pi}:\mathcal{O}\rightarrow\mathcal{A}$.
In continuous domains it additionally depends on the timestep $t \in \mathbb{N}$ from the start
of the episode, $\hat{\pi}:\mathcal{O}\times\mathbb{N}\rightarrow\mathcal{A}$, and we factor it
into a synthesized \emph{feature map} $\phi:\mathcal{O}\times\mathbb{N}\rightarrow\mathbb{R}^{m}$,
whose $m$ components are short programs computing interpretable quantities of the observation and
of $t$, and a fitted matrix $W\in\mathbb{R}^{|\mathcal{A}|\times m}$:
\begin{equation}
\hat{\pi}(o,t) \;=\; \text{clip}\!\left(W\,\hat{\phi}(o,t),\; a_{\min},\; a_{\max}\right).
\label{eq:formfit}
\end{equation}
The two operations in Equation~\ref{eq:formfit} serve different purposes. The features are
normalized by a running mean and variance, written $\hat{\phi}$, because they are not on a
common scale: a constant bias contributes $1.0$, a periodic term lies in $[-1,1]$, and a clipped
velocity may span an order of magnitude more. The search perturbs every weight by the same fixed amount. So without normalization, equal weight perturbations could induce very different changes in the controller output depending on the scale of the corresponding feature. The outer clip bounds the result to the environment's admissible torque range $[a_{\min}, a_{\max}]$, which the unbounded product $W\hat{\phi}$ might not respect.

The timestep provides an explicit temporal signal to the controller. Making
$\phi$ a function of $t$ allows the LLM to synthesize periodic features such as
$\sin(\omega t)$ and $\cos(\omega t)$, whose phase can coordinate actuator
torques over a gait cycle. The controller can therefore combine reactive
feedback from the current observation with a recurring control rhythm. This does 
not fully depend on trajectory history since $t$ records only elapsed time, but
it provides a compact phase signal that is useful for rhythmic locomotion.
Running a policy for one episode yields a trajectory $\tau$ and return $r$.
The agent is given an episode budget $N$ in discrete domains, and in continuous
domains a budget of $L$ feature maps together with a per-map budget of search
episodes.

\subsubsection{Controller Synthesis and Parameter Optimization}

The two settings differ in which component of the controller is improved
through interaction. The agent begins from an initial observation $o_0$ and
prompt context
$c=\langle \text{state schema},\mathcal{A},\text{object API},\mu,o_0\rangle$,
where $\mathcal{A}$ is the primitive action space and $\mu$ is a generic
mission statement. The LLM is given neither the environment dynamics nor a
demonstration.

\paragraph{Discrete domains.}
The LLM first synthesizes an executable policy
$\pi_0:\mathcal{O}\rightarrow\mathcal{A}$ from the prompt context $c$.
Executing the policy for an episode produces a trajectory $\tau_i$ and return
$r_i$. On each subsequent episode, the LLM is conditioned on
$(c,\pi_i,\tau_i,r_i)$ and synthesizes a revised policy $\pi_{i+1}$.
Revision is always applied to the most recent policy, while the best controller
encountered over the run is retained separately.

\paragraph{Continuous domains.}
We first collect a set of transitions $\mathcal{D}_{\mathrm{rand}}$ using
random actions. The LLM uses $(c,\mathcal{D}_{\mathrm{rand}})$ to synthesize
the programmatic structure of the controller as a feature map
$\phi_i:\mathcal{O}\times\mathbb{N}\rightarrow\mathbb{R}^m$.
For each candidate feature map:
\begin{enumerate}
    \item An LLM synthesizes $\phi_i$ from
    $(c,\mathcal{D}_{\mathrm{rand}})$.
    \item A derivative-free optimizer fits the numerical parameters $W_i$
    against measured episode returns. 
    \item The resulting controller is evaluated in the environment.
\end{enumerate}

For a fixed feature map $\phi_i$, the parameter-fitting objective is $W_i^{*}
=
\arg\max_W
\mathbb{E}_{\tau \sim \pi_{W,\phi_i}}
\left[r(\tau)\right]$,
where $\pi_{W,\phi_i}$ denotes the controller in
Equation~\ref{eq:formfit} with feature map $\phi_i$ fixed.
We optimize this objective using derivative-free search. In the reported
experiments, we use Augmented Random Search \citep[ARS;][]{mania2018simple}. 
We repeat the procedure for $L$ independently synthesized feature maps and
retain the best fitted controller. Unlike in the discrete setting, the
programmatic feature map is not revised from its achieved return. After
$\phi_i$ is synthesized, return feedback is used only to fit the numerical
parameters $W_i$ for that candidate.
Algorithm~\ref{alg:controller} summarizes the full procedure.

\section{Experiments}

Our experiments seek to answer the following questions:
(1) Can programmatic policy synthesis effectively control complex environments while avoiding the computational cost of planning through a learned world model at every decision?
(2) Do the resulting programmatic policies remain effective when the environment dynamics change?
(3) Can structured programmatic policy synthesis scale to continuous-control domains by combining LLM-synthesized policy structure with numerical optimization?

\subsection{Experimental Setup}

\paragraph{Domains and Observations.}
We evaluate our approach across Atari games, Flappy Bird, and MuJoCo continuous-control environments \citep{todorov2012mujoco}. For Atari, we use OCAtari \citep{delfosse2024ocatari} to parse each image frame into a list of objects, each with an object category, bounding box, and velocity. Flappy Bird exposes the same object-level schema. The agent acts using the environment's primitive action space.

\subsection{Learning Programmatic Policies in Atari}
\label{sec:atari}

\paragraph{Baselines.}
For the LLM-based baselines, we compare against ReAct \citep{yao2023react}, which uses an LLM directly as a policy to select actions, and WorldCoder \citep{tang2024worldcoder}, which synthesizes a Python transition model from experience and plans through the learned model. We also compare against PPO \citep{schulman2017proximal}, a standard model-free reinforcement learning baseline. Finally, on Pong we compare against PoE-World \citep{piriyakulkij2026poe}, which represents the transition model as a composition of programmatic experts and uses the learned model for planning. We restrict the PoE-World comparison to Pong because it requires human gameplay demonstration, which is only available for Pong among the games in our evaluation suite.

\paragraph{Evaluation.}
We report the best episode score within a run, taking the median over three seeds per game, together with the costs of obtaining that controller: the LLM calls issued, the tokens consumed, and the environment steps taken before the episode in which it was found. Our method issues one call per episode and completes within $20$. ReAct issues one call per action, so a call cap is incompatible with running it at all. We cap it by tokens instead. WorldCoder's search over candidate transition models is given $30$ calls,
which suffices for it to return a model. PoE-World is additionally given a $650$-step human
demonstration that no other method receives.

We additionally report execution cost as decision latency: the time to evaluate the synthesized policy on a single observation. This is measured over $300$ states encountered during play and reported as the median across states. For methods that deliberate at each step, the corresponding quantity is the time to produce one action. For ReAct this is a language model call, whereas for planning-based methods it
is planning time divided by the actions that planning supplied.

\begin{figure}[t]
\centering

\begin{minipage}[t]{0.70\textwidth}
{\raggedright{\sffamily\bfseries\large A}\par}\vspace{2pt}
\centering
\scriptsize
\setlength{\tabcolsep}{2.5pt}
\begin{tabular}{lrrrrrr}
\toprule
Game & \textbf{\shortstack[r]{Code to\\Control}} & \shortstack[r]{PPO\\@100k} & \shortstack[r]{PPO\\@20M} & ReAct & \shortstack[r]{World-\\Coder} & Random \\[2pt]
\midrule
Pong  & $\mathbf{+18}$     & $-21$ & $+21$     & $-17$ & $-19$ & $-21$ \\
Space Invaders & $\mathbf{490}$     & $105$ & $115$     & $105$ & $310$ & $42$  \\
Asterix        & $\mathbf{3{,}100}$ & $150$ & $1{,}200$ & $750$ & $600$ & $175$ \\
Breakout       & $\mathbf{19}$      & $1$   & $21$      & $3$   & $2$   & $1$   \\
Fishing Derby  & $\mathbf{+31}$     & $-99$ & $-95$     & $15$  & $-60$ & $-96$ \\
Freeway        & $\mathbf{22}$      & $21$  & $24$      & $19$  & $0$   & $0$   \\
\bottomrule
\end{tabular}
\end{minipage}\hfill
\begin{minipage}[t]{0.285\textwidth}
{\raggedright{\sffamily\bfseries\large B}\par}\vspace{2pt}
\centering
\scriptsize
\setlength{\tabcolsep}{2.5pt}
{\begin{tabular}{crr}
\toprule
\textbf{\shortstack{Code to\\Control}} & ReAct & \shortstack[r]{World-\\Coder} \\[2pt]
\midrule
  82k & 5.0M & 70k \\
  92k & 4.5M & 395k \\
  89k & 5.3M & 142k \\
  76k & 1.6M & 129k \\
  77k & 20.3M & 153k \\
  71k & 25.9M & 167k \\
\midrule
  \textbf{487k} & 62.5M & 1.1M \\
\bottomrule
\end{tabular}}
\end{minipage}


\begin{minipage}[t]{0.585\textwidth}
{\raggedright{\sffamily\bfseries\large C}\par}\vspace{2pt}
\centering
\scriptsize
\setlength{\tabcolsep}{3pt}
\begin{tabular}{lrrrr}
\toprule
Method & Score & LLM calls & Tokens & Env.\ steps \\
\midrule
\textbf{Code to Control} & $\mathbf{+18}$ & \textbf{7} & \textbf{27k} & \textbf{25{,}312} \\
ReAct                               & $-17$ & $3{,}364$ & 5.0M & $3{,}292$ \\
PoE-World                           & $-17$ & $1{,}688$ & 1.5M & $3{,}714 + 650^{\S}$ \\
WorldCoder                          & $-19$ & $30$      & {70k}  & $8{,}678$ \\
PPO @ 100k                          & $-21$ & ---       & ---  & $100{,}000$ \\
PPO @ 20M                           & $+21$ & ---       & ---  & $20{,}000{,}000$ \\
Random                              & $-21$ & ---       & ---  & --- \\
\bottomrule
\end{tabular}
\end{minipage}\hfill
\begin{minipage}[t]{0.395\textwidth}
{\raggedright{\sffamily\bfseries\large D}\par}\vspace{2pt}
\centering
{\includegraphics[width=\linewidth]{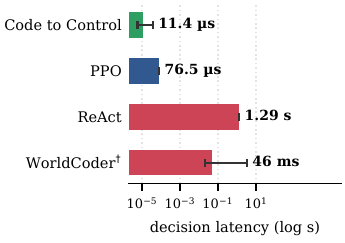}}
\end{minipage}
\caption{\textbf{Atari evaluation.}
\textbf{A}: Median best score over three seeds.
\textbf{B}: Total LLM tokens.
\textbf{C}: Pong performance and interaction cost up to the reported controller; PoE-World additionally uses a 650-step human demonstration.
\textbf{D}: Median decision latency on a log scale, with 95\% confidence intervals.
$^{\dagger}$WorldCoder's planning time is divided by the actions of the episode in which
it planned. Its error bar is the range of its three runs.}
\label{fig:atari}
\end{figure}

\paragraph{Performance and interaction efficiency.}
Code to Control scores higher than WorldCoder and ReAct on all six Atari
games (Figure~\ref{fig:atari}\textbf{A}), and also exceeds PPO trained for
100k environment steps on all six. On Pong, where we additionally compare
against PoE-World, Code to Control reaches a score of $+18$, compared with
$-17$ for both ReAct and PoE-World and $-19$ for WorldCoder, making it the
only LLM-based method in our comparison to achieve a positive score.

Even when PPO is trained for 20 million environment steps, Code to Control
remains ahead on Space Invaders, Asterix, and Fishing Derby. PPO achieves
higher scores on Pong, Breakout, and Freeway, but only after orders of
magnitude more environment interaction. Thus, substantially increasing PPO's
interaction budget closes the gap on some games, but Code to Control remains
ahead on half of the suite despite using far fewer environment interactions.

The interaction cost of Code to Control is comparatively small. Across the
six games, our method uses 120 LLM calls and 487k tokens in total
(Figure~\ref{fig:atari}\textbf{B}). On Pong, the final controller is found
after 7 LLM calls and 25,312 environment steps
(Figure~\ref{fig:atari}\textbf{C}), compared with the 20 million environment
steps used by the strongest PPO baseline.

\paragraph{Decision-time efficiency.}
The efficiency advantage also extends to action selection after learning.
Under our measurement protocol, Code to Control selects an action in
$11.4\,\mu$s, compared with $76.5\,\mu$s for PPO and $1.29$\,s for ReAct.
Meanwhile, WorldCoder's planning costs $46$\,ms per action
(Figure~\ref{fig:atari}\textbf{D}). Thus, the synthesized controller is
faster to execute than the PPO policy forward pass in this setting, while
avoiding the much larger per-decision cost of LLM inference or planning.

\paragraph{Analysis.}
To understand what the LLM learns through interaction, we inspect the
programs produced across the Atari runs. Many champion controllers use object
velocities to anticipate future states rather than reacting only to current
positions. Successful revisions tend to introduce small, concrete changes to
the existing controller, such as predicting where a moving object will be in
the near future, constraining predicted positions to the playing field,
adding a tolerance region around a target to avoid unnecessary oscillation,
or handling cases in which tracked objects disappear. In Space Invaders, for
example, revisions introduce explicit threat detection for incoming
projectiles, estimating the time to impact of bullets approaching the player
and taking evasive actions.

Pong provides a particularly clear example of this process. A
negative-scoring controller steered the paddle toward the ball's current
position, \texttt{target\_y = ball.y}, and held still whenever the paddle was
within $8$ pixels of that target. The resulting trajectory showed the paddle
repeatedly arriving late to the ball. After observing this trajectory, the
LLM revised the target to extrapolate the ball's future position using its
velocity,
\texttt{predicted\_ball\_y = ball.y + ball.dy * lead\_ticks}, with
\texttt{lead\_ticks = 5}, and reduced the tolerance region from $8$ to $4$
pixels. The score improved from $-8$ to $+18$ in the following episode, and
the revised program is the champion reported for Pong
(Box~\ref{box:champ-pong}). More broadly, these examples suggest that
interaction typically adds task-specific control structure over successive
revisions rather than replacing the controller wholesale. Complete examples
of synthesized controllers are provided in Appendix~\ref{app:programs}.

\paragraph{Search dynamics and failure modes.}
Policy revision is not monotonic. Across the 360 Atari episodes in our
evaluation, 72\% score below the best controller found so far within their
run, and 28\% decrease by more than 25\% relative to the immediately
preceding episode. Consequently, retaining the best controller encountered
during synthesis is important: a useful revision can be followed by a worse
one even when later revisions continue from the most recent program.

The champion controller is found at a median episode index of 5 out of 20,
although in some runs improvement continues as late as episode 19. The main
failure mode is therefore not invalid program generation, but revisions that
reduce controller performance. Across these runs, every synthesized revision
parsed, no episode returned an invalid action, and no controller crashed
during execution. Instead, later revisions sometimes replace effective
control logic with a worse alternative, while champion banking preserves the
strongest program found earlier in the run.

\subsection{Robustness to Changes in Environment Dynamics}
\label{sec:flappy}

Flappy Bird tests whether a synthesized controller remains useful when the
environment dynamics change. We synthesize controllers only on the base game,
using the same object-level observation schema as Atari and a budget of 20
LLM calls. We then evaluate the resulting controllers under eight changes to
gravity, drag, flap strength, pipe speed, and gap width. Each controller is
first evaluated unchanged. If it fails before the 20,000-step evaluation cap,
we hold the synthesized program fixed and refit only its numerical constants
using feedback from the modified environment, with no additional LLM calls
or program synthesis.

\begin{figure}[t]
\centering

\begin{minipage}[t]{0.52\textwidth}
{\raggedright{\sffamily\bfseries\large A}\par}\vspace{2pt}
{\centering
\scriptsize
\setlength{\tabcolsep}{2pt}
\resizebox{\linewidth}{!}{
\begin{tabular}{lccrrr}
\toprule
System & \shortstack{Survives\\the cap} & Score &
\shortstack[r]{Env.\\steps} & Calls & Tokens \\[2pt]
\midrule
\textbf{Code to Control} & \textbf{yes} & 311$^{*}$ &
\textbf{21{,}419} & \textbf{20} & \textbf{96k} \\
ReAct          & no & 5 & 356 & 314 & 44.5M \\
WorldCoder     & no & 1 & 514 & 30  & 191k \\
Random / no-op & no & 0 & --- & --- & --- \\
\bottomrule
\end{tabular}
}\par}
\vspace{6pt}
\caption{\textbf{Flappy Bird evaluation.}
\textbf{A}: Base-game performance; $^{*}$score is cap-limited.
\textbf{B}: Survival under eight dynamics changes, zero-shot (open) and
after parameter refitting (filled).}
\label{fig:flappy}
\end{minipage}\hfill
\begin{minipage}[t]{0.46\textwidth}
{\raggedright{\sffamily\bfseries\large B}\par}\vspace{2pt}
\centering
\includegraphics[width=\linewidth]{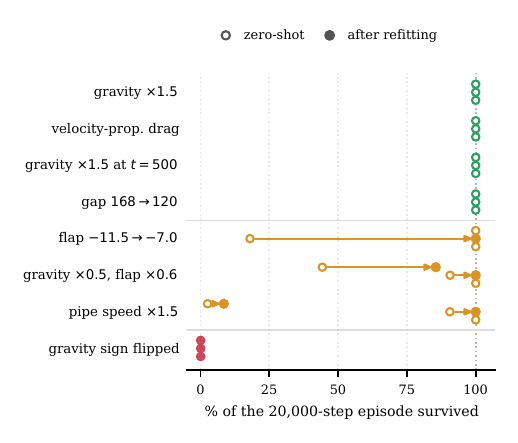}
\end{minipage}
\end{figure}

\paragraph{Base-game performance.}
Code to Control learns a controller that survives the full 20,000-step
evaluation cap (Figure~\ref{fig:flappy}\textbf{A}), using 21,419 environment
steps, 20 LLM calls, and 96k tokens. Because the underlying environment has
no episode limit and pipes arrive at a fixed rate, the score of a controller
that survives the entire evaluation is determined by the imposed cap. We
therefore treat survival as the primary measure of successful play. ReAct and
WorldCoder terminate substantially earlier, reaching scores of 5 and 1,
respectively. PPO reaches comparable sustained play after 2 million
environment steps, roughly $93\times$ the interaction used by Code to
Control.

\paragraph{Transfer and parameter refitting.}
The synthesized control structure transfers across substantial changes in
the environment dynamics (Figure~\ref{fig:flappy}\textbf{B}). All three
champions survive the full evaluation cap without adaptation under increased
gravity, velocity-dependent drag, a mid-episode increase in gravity, and a
narrower pipe gap. For the cells where the frozen controller fails, refitting only its numerical constants restores full survival in three of the five and improves the other two without reaching the cap. This is done without changing the program or calling the LLM again. Reversing the sign of gravity is the exception. Parameter refitting does not recover successful play since that would require a structural program change. These results indicate that the synthesized program can capture control structure that remains useful across changes in physics, while its numerical parameters can be adapted separately when necessary.

\subsection{Scaling to Continuous Control}
\label{sec:continuous}

We next evaluate whether Code to Control can scale to continuous-control
domains in MuJoCo \citep{todorov2012mujoco}. Unlike Atari and Flappy Bird,
where a synthesized program can directly return a discrete action, these
environments require producing a numerical command for each actuator. We
therefore separate the controller into two components: the LLM synthesizes a
programmatic feature map $\phi$, and derivative-free optimization fits the
matrix $W$ that maps those features to continuous actions. Our goal in this setting is to test whether Code to Control can produce
effective continuous controllers. We do not make a matched
environment interaction-efficiency comparison in MuJoCo.

\paragraph{Baselines.}
We compare against GIF-MCTS \citep{dainese2024generating} and WorldCoder
\citep{tang2024worldcoder}, which synthesize transition models and act by
planning through them. Both are run from the authors' unmodified
implementations using the same LLM backend and sampling settings, with their
published budget of 30 LLM calls per environment. The MuJoCo physics engine
is unavailable during model synthesis, preventing either method from calling
the true simulator.

We additionally report the published CQL and oracle-planning results from
\citet{dainese2024generating}. The oracle replaces the learned transition
model with the true MuJoCo simulator \citep{todorov2012mujoco} while
retaining the same planning procedure, isolating model error from limitations
of the planner itself.

\paragraph{Performance.}
Code to Control scores higher than both planning-based program synthesis
baselines on nine of the ten MuJoCo environments
(Figure~\ref{fig:mujoco}). The differences are especially large on
several locomotion tasks. On HalfCheetah, Code to Control reaches
$3{,}958.95$, while GIF-MCTS and WorldCoder obtain $-87.4$ and $-69.4$,
respectively. On HumanoidStandup, it reaches $127{,}370$, approximately
$4.3\times$ the return of either planning baseline. The one exception is
Reacher, where both planning baselines obtain slightly higher returns.
Code to Control also exceeds the published CQL result on all ten environments.

\paragraph{Oracle comparison.}
On six of the ten environments, Code to Control also scores higher than the
oracle planner reported by \citet{dainese2024generating}, despite the oracle
having access to the true MuJoCo simulator. Because this comparison removes
transition-model error while retaining the same planning procedure, the
result suggests that model accuracy alone does not explain the gap on these
tasks; the planning procedure itself can also limit performance.

\paragraph{Controller structure and locomotion.}
The strong returns on HalfCheetah and Swimmer reflect sustained forward
motion rather than a per-step survival reward. Neither environment provides
an alive bonus: reward is determined by forward velocity minus control cost.
The reported HalfCheetah controller travels $212.55$\,m over 1,000 steps,
with a mean forward velocity of $4.25$\,m/s, while the reported Swimmer
controller travels $12.83$\,m with a mean forward velocity of $0.321$\,m/s.

The synthesized control terms combine several intuitive forms of feedback.
In HalfCheetah, the LLM proposes terms describing body posture, leg
configuration, forward motion, limb asymmetry, and periodic timing signals.
After fitting, the larger weights are assigned mainly to terms that depend on
the current physical state, while the periodic terms receive smaller weights.
The Swimmer controller shows a similar pattern, relying strongly on current
motion and joint state rather than on timing signals alone. Thus, the fitted
controllers do not rely primarily on a fixed periodic gait. They use the
current physical state to continuously adjust their actions.

\paragraph{Interpreting MuJoCo returns.}
Episode return does not always correspond directly to forward locomotion in
MuJoCo because the reward functions differ across tasks. Hopper, Walker2d,
Ant, and Humanoid include a per-step alive reward, so a controller can obtain
substantial return while moving very little. For example, the reported
Humanoid controller has a median displacement of $-0.44$\,m despite a return
of $472.92$, with $512$ of accumulated alive reward. The reported Hopper and
Walker2d controllers move only $1.46$\,m and $3.12$\,m, respectively, over
1,000 steps. By contrast, HalfCheetah and Swimmer have no alive reward. Their
reported controllers travel $212.55$\,m and $12.83$\,m, with mean forward
velocities of $4.25$\,m/s and $0.321$\,m/s, respectively.

\begin{figure}[t]
\centering
\begin{minipage}{\textwidth}
\centering
\footnotesize
  \setlength{\tabcolsep}{3pt}
\begin{tabular}{lrrrrr}
\toprule
Domain & \textbf{Code to Control} & GIF-MCTS & WorldCoder & CQL & \textit{Oracle} \\
\midrule
Hopper & \textbf{1{,}155.40} & 56.6 & 8.1 & 137.4 & 229.1 \\
HalfCheetah & \textbf{3{,}958.95} & $-$87.4 & $-$69.4 & $-$1.3 & 893.3 \\
Walker2d & \textbf{1{,}388.54} & 31.9 & 8.5 & 278.0 & 334.7 \\
Swimmer & \textbf{320.53} & $-$11.2 & $-$18.7 & 28.4 & 317.8 \\
Humanoid & \textbf{472.92} & 153.9 & 299.2 & 393.3 & 1{,}860.7 \\
Ant & \textbf{1{,}003.93} & 448.4 & 396.6 & 998.0 & 1{,}304.7 \\
HumanoidStandup & \textbf{127{,}370} & 29{,}715 & 29{,}671 & 51{,}046 & 138{,}076 \\
Inverted Pendulum & \textbf{1{,}000.00} & 4.6 & 15.4 & 66.7 & 42.5 \\
Inv.\ Double Pendulum & \textbf{9{,}358.85} & 84.4 & 99.8 & 164.0 & 241.6 \\
Reacher & $-$9.24 & \textbf{$-$7.0} & $-$8.9 & $-$11.5 & $-$6.8 \\
\bottomrule
\end{tabular}
\end{minipage}
\caption{\textbf{MuJoCo continuous-control evaluation.}
Episode return, reported as the median over independent runs.
GIF-MCTS and WorldCoder use their published budget of 30 LLM calls per
environment, compared with 6 for Code to Control.
Further evaluation details are described in Section~\ref{sec:continuous}.}
\label{fig:mujoco}
\end{figure}

\section{Related Work}

\paragraph{Theory-based RL and code world models.}
Theory-based reinforcement learning learns structured models of environment dynamics and uses them for planning
\citep{tsividis2026human, ahmed2025synthesizing, ahmed2026learning}.
Recent LLM-based systems instantiate such models as executable code. WorldCoder synthesizes Python transition and reward models and plans through
them \citep{tang2024worldcoder}. GIF-MCTS searches over code-based world models and uses a separate planner for action selection, including continuous
control in MuJoCo \citep{dainese2024generating}. PoE-World composes smaller
programmatic experts into a probabilistic world model
\citep{piriyakulkij2026poe}. It additionally uses a gameplay demonstration that shows causal interactions in the environment and assumes access to the
task goal. Code to Control requires neither, collecting its own interaction data and learning the goal while synthesizing a programmatic policy. The resulting controller selects actions without simulating a learned world model or invoking a planner at
decision time.

\paragraph{Programmatic policies.}
Programmatic Interpretable Reinforcement Learning \citep[PIRL;][]{verma2018programmatically} represents policies as executable programs, often within a domain-specific language (DSL). PIRL first learns a neural policy and then uses it to direct local search over programmatic
policies. PROPEL alternates policy-gradient updates in a larger policy space with projection back into a programmatic
policy class through program synthesis and imitation learning \citep{verma2019imitation}. LEAPS learns a continuous embedding of programs and searches that space for programs that maximize task reward \citep{trivedi2021learning}. More recently, LLM-GS uses an LLM to generate policy code, converts it into a DSL, and further improves the resulting program using Scheduled Hill Climbing \citep{liu2025synthesizing}. These
approaches ultimately constrain the learned policy to a predefined DSL or symbolic program space. Code to Control instead synthesizes general-purpose Python policies that execute directly, without translation into a DSL, and
learns them through interaction in Atari and continuous-control domains.

\paragraph{LLMs as policies and as code generators for control.}
LLMs have also been used directly for action selection and control. In ReAct, the LLM itself acts as the policy, producing actions during interaction with the environment \citep{yao2023react}. Code as Policies uses an LLM to translate natural-language commands into executable robot policy code that composes provided perception and control-primitive APIs \citep{liang2023code}. Voyager similarly uses LLMs to construct and reuse executable skills in Minecraft \citep{wang2023voyager}, while Eureka uses
an LLM to synthesize reward functions and leaves policy optimization to reinforcement learning \citep{ma2024eureka}. In contrast, Code to Control
learns from interaction and reward rather than natural-language task commands, and does not assume a predefined task-specific control-primitive library. The synthesized controller acts directly in the environment's
primitive action space and controllers are constructed with tunable parameters that support adaptation when new dynamics are encountered. 

\paragraph{Derivative-free optimization for continuous control.}
Augmented Random Search (ARS) is a derivative-free optimization method that searches policy parameters through random perturbations and has been shown to
perform competitively with reinforcement-learning methods on MuJoCo locomotion tasks \citep{mania2018simple}. Evolution strategies similarly
optimize policies through parameter-space search without backpropagating
through the environment \citep{salimans2017evolution}. Code to Control uses
this style of optimization differently. An LLM first synthesizes the programmatic feature representation, and ARS then fits only the numerical matrix mapping those features to actions. This separates synthesis of the
controller's structure from optimization of its numerical parameters.

\section{Discussion}

Code to Control demonstrates that program synthesis can be used to efficiently learn
fast controllers that execute directly in the environment. Rather than
invoking an LLM or searching through a learned world model at every decision,
the resulting policy is ordinary executable code (in Python). This shifts computation
from decision time to policy synthesis, enabling real-time control while
retaining the structured and interpretable form of a program.

\paragraph{Tradeoff with world models.}
Our experiments focus on reactive control problems where effective behavior can be produced directly from the current observation, making repeated
multi-step planning at every decision less compelling. In these settings,
synthesizing the controller itself shifts computation away from decision
time. Once learning is complete, actions require only execution of the
controller, with no model rollout, planning procedure, or LLM call. In
domains where explicit prediction or long-horizon lookahead is important,
planning-based world models may be preferable.

\paragraph{Limitations.}
The present formulation leaves several aspects of the control problem
outside its scope. Code to Control currently operates on structured state
representations rather than learning directly from pixels. Furthermore, allowing
the synthesized program to change can improve behavior, but excessive
revision can also destroy useful structure, while fixing the structure can
limit adaptation. Identifying an effective operating point along this
tradeoff remains open. Performance is also shaped by the objective supplied
by the environment. In several MuJoCo tasks, for example, substantial return
can be obtained from survival with little locomotion. 

\subsection*{AI use statement}


Generative AI tools were used to support literature retrieval and discovery,
assist with drafting and polishing writing, and support research ideation and
execution. All AI-assisted work was reviewed and verified by the authors, who
take responsibility for the final content of the paper.




\subsection*{Reproducibility statement}


Code is available at \url{https://github.com/ZerghamAhmed/code-to-control}. The full
algorithm is given in Appendix~\ref{app:algorithm}, the prompts used are in Appendix~\ref{app:prompts}, and example synthesized controllers
in Appendix~\ref{app:programs}. Atari and Flappy Bird experiments use the
language model \texttt{claude-haiku-4-5}. MuJoCo experiments use
\texttt{claude-opus-4-8}. Within each domain, every method uses the same model.


\subsubsection*{Acknowledgments}
SJG is supported by the Kempner Institute for the Study of Natural and Artificial Intelligence, a Polymath Award from Schmidt Sciences, and the Department of Defense MURI program under ARO grant W911NF-2310277.

\bibliography{iclr2027_conference}
\bibliographystyle{iclr2027_conference}

\appendix
\section{Appendix}

\subsection{Code to Control Algorithm}
\label{app:algorithm}

\begin{algorithm}[!t]
\caption{Code to Control}
\label{alg:controller}
\begin{algorithmic}[1]
\Require environment $E$, LLM, initial observation $o_0$, action space $\mathcal{A}$,
generic mission $\mu$, episode budget $N$, feature-map budget $L$,
number of feature terms $m$
\Ensure controller $\pi$

\State $c \gets \langle
\text{state schema},\ \mathcal{A},\ \text{object API},\ \mu,\ o_0
\rangle$
\Comment{Prompt context; no goal or dynamics given}

\If{$\mathcal{A}$ is discrete}
    \State $\pi_1 \gets \text{LLM}(c)$
    \Comment{Synthesize executable policy}

    \For{$i = 1$ \textbf{to} $N$}
        \State $\tau_i, r_i \gets \text{Rollout}(\pi_i, E)$
        \Comment{Trajectory and return of episode $i$}

        \State $\text{Bank}(\pi_i, r_i, |\tau_i|)$
        \Comment{Retain best controller encountered}

        \If{$i < N$}
            \State $\pi_{i+1} \gets
            \text{LLM}(c, \pi_i, \tau_i, r_i)$
            \Comment{Revise latest policy from its own play}
        \EndIf
    \EndFor

\Else
    \State $\mathcal{D}_{\mathrm{rand}} \gets$
    Generate transitions with random actions

    \For{$i = 1$ \textbf{to} $L$}
        \State $\phi_i \gets
        \text{LLM}(c, \mathcal{D}_{\mathrm{rand}}, m)$
        \Comment{Independent feature-map proposal}

        \State $W_i \gets
        \textsc{OptimizeWeights}(\phi_i, E)$
        \Comment{Derivative-free optimization against returns}

        \State $\pi_i \gets (o,t) \mapsto
        \text{clip}\!\left(
        W_i\hat{\phi}_i(o,t),
        a_{\min},
        a_{\max}
        \right)$
        \Comment{$\hat{\phi}_i$: Normalize $\phi_i$ online}

        \State $r_i \gets \text{Evaluate}(\pi_i, E)$
        \State $\text{Bank}(\pi_i, r_i)$
    \EndFor
\EndIf

\State \textbf{return} banked champion
\end{algorithmic}
\end{algorithm}

\subsection{Language model prompts}
\label{app:prompts}

The two prompts below are the ones used in the reported runs, reproduced verbatim from
the files.

The prompts reflect different uses of the same framework. Code to Control
separates program structure from numerical parameters, with parameter fitting
used when useful for the task. For Atari and the initial Flappy Bird controller,
the policy can straightforwardly return a discrete action. We therefore use the
simpler instantiation, where the LLM synthesizes and revises the complete
executable controller, including its numerical constants.

In the Flappy Bird adaptation experiments, parameter fitting instead serves as
an adaptation mechanism. It allows the controller to adjust to changes in
environment dynamics while keeping the synthesized program structure fixed. In
MuJoCo, where the controller must produce continuous actuator commands,
separating program structure from numerical parameters is beneficial. The LLM
synthesizes the feature map, while derivative-free search fits the numerical
parameters mapping those features to actions.

\begin{itemize}
\item Box~\ref{box:prompt-discrete}: the discrete-action controller prompt for
every Atari game and for Flappy Bird.
\item Box~\ref{box:prompt-continuous}: the continuous prompt, which asks for a library of
candidate terms rather than a controller.
\end{itemize}

\begin{mybox}[boxrule=0pt, parbox=false, opacityframe=1, colframe=black, label=box:prompt-discrete]{Discrete controller synthesis and revision (Atari and Flappy Bird)}
\begin{lstlisting}[basicstyle=\ttfamily\scriptsize, breaklines=true, columns=fullflexible, keepspaces=true]
You are building and refining a game-playing agent's CONTROLLER.
Everything is Python. There is NO PDDL, NO world model, and NO planning.

You are given, and must return, ONE thing:

A) A REACTIVE CONTROLLER. `policy(state) -> action`, mapping the current state
   directly to the next primitive action. No search, no simulation, no lookahead.

You do NOT write a transition model, a plan, operators, or a heuristic. Nothing
simulates the future. Whatever your policy returns is played immediately.

STATE SCHEMA (identical convention in every game):
- positions are box CENTERS in raw pixels; y increases DOWNWARD.
- `<entity>` -> [[x, y], ...] (a LIST - there may be several instances);
  `<entity>_size` -> [[width, height], ...]; `<entity>_velocity` -> per-tick
  position deltas [[dx, dy], ...].
- some games also expose `<entity>_orientation` -> [h, ...]: a small integer
  heading per instance. When present, this field explicitly records the entity's
  heading, including rotations that may occur without any change in position or
  velocity; correlate it with observed transitions to infer the relevant geometry.
- plus `score` [n], `won`, `lost`.
- objects may APPEAR or DISAPPEAR between ticks (entering/leaving the
  observable world) - handle absent keys/instances gracefully.

An OBJECT API is PROVIDED in your namespace - prefer it over raw coordinate math.
It is ALREADY IN SCOPE: call `objects(state)` directly. Do NOT import it. There is
no module to import it from, and `from objects import objects` inside `policy`
compiles and leaves `policy` callable, so nothing looks wrong until it runs - then
every tick raises ModuleNotFoundError, the runner substitutes its fallback action,
and the whole episode is lost. `import math` is fine; nothing else is needed.

__OBJAPI__

HOW THE RUNNER EXECUTES WHAT YOU RETURN - mechanism, so you can reason about it:
- Every tick it calls `policy(state)` with the CURRENT REAL state and plays the
  action you return. There is no plan to cycle and no horizon to run out.
- Your policy therefore sees ground truth every tick and never accumulates
  prediction error. It also gets no chance to look ahead: any anticipation must
  be written into the policy itself, from the current state.
- If `policy` raises, or returns something not in the primitive action list, the
  runner takes a fallback action and records the failure as evidence. A policy
  that crashes on an unusual state is worse than a crude one that never does.

WHAT MAKES A GOOD CONTROLLER HERE:
- React to the RELATIVE geometry that decides the next few ticks (distances and
  velocity differences between the player and the things that matter), not to
  absolute coordinates.
- Anticipate. Because there is no lookahead, a policy that acts on where things
  ARE will act too late; one that acts on where they are GOING will not. Velocity
  fields, or differences between successive positions, are how you do that.
- Prefer a small number of clear cases over one elaborate expression. Threshold
  and sign logic is legitimate and often optimal when the action set is discrete.
- Numeric constants are YOUR responsibility in this run - nothing fits them for
  you. Pick values you can justify from the observed transitions below, and treat
  a constant that the evidence does not pin down as a thing to revise next round.

HOW TO USE THE OBSERVED TRANSITIONS BELOW
They are ground truth from real play under your PREVIOUS controller. They tell you
what your policy actually did, what it scored, and where it failed. Read the ticks
immediately BEFORE a death: that is where the policy chose wrong, and it is almost
always a case your current conditions do not cover.

CURRENT CONTROLLER:

__ABSTRACTIONS__

OBSERVED TRANSITIONS (ground truth from your previous controller's play):
__TRAJECTORY__

PRIMITIVE ACTIONS available this game: __ACTIONS__

CURRENT RAW STATE (your policy must handle exactly these keys):
__RAW_STATE__

MISSION:
__MISSION__

OUTPUT - return the following fenced block and nothing else:

1) the updated controller:
```python
# controller.py
import math

def policy(state):
    # return ONE action from the primitive action list above
    ...
```
\end{lstlisting}
\end{mybox}

Only the first of the sampled transitions is shown below. They make up most of the
prompt's length and a fresh sample is drawn for each library.

\begin{mybox}[boxrule=0pt, parbox=false, opacityframe=1, colframe=black, label=box:prompt-continuous]{Continuous feature-map synthesis (MuJoCo)}
\begin{lstlisting}[basicstyle=\ttfamily\scriptsize, breaklines=true, columns=fullflexible, keepspaces=true]
You are proposing a LIBRARY of candidate control terms for an unknown dynamical
system. You are NOT choosing which ones matter -- a search will fit a weight for every term and
drive the useless ones to zero. So propose LIBERALLY and DIVERSELY: include terms you are unsure
about. A term that turns out to be useless costs nothing; a term you failed to propose can never
be used.

Return exactly this shape:

```python
def terms(state, t):
    # state: dict, numeric fields are 1-element lists, e.g. state['rootz'][0]
    # t: integer tick counter, starting at 0
    # return a LIST of 14 floats -- the value of each candidate term right now.
    # Each will be multiplied by a fitted weight and summed into every actuator.
    ...
```

WHAT MAKES A GOOD LIBRARY
* DIVERSE MECHANISMS, not variations of one. Include, for example: raw state fields; velocities;
  a periodic term (and a second one at a different frequency, or shifted in phase); products or
  differences of two fields; a constant bias; something that grows or saturates.
* Terms should be O(1) in magnitude. A term that is always ~1000 will dominate before fitting
  starts. Normalise or clip where sensible.
* Return exactly 14 floats for EVERY state, finite always. Guard missing keys. Only `math` may
  be imported. Module-level state between calls is allowed.
* Do NOT try to solve the task in one term. Each term is a BUILDING BLOCK; the fit composes them.

Reply with a single ```python block containing `terms`, nothing else.

## Real transitions (state, action, next_state, reward earned on that step)
{"state": {"rootz": -0.0911, "rooty": 0.0815, "bthigh": -0.0675, "bshin": 0.053, "bfoot": -0.0004, "fthigh": 0.0616, "fshin": -0.0616, "ffoot": 0.0014, "rootx_velocity": 0.0724, "rootz_velocity": 0.1594, "rooty_angular_velocity": -0.1077, "bthigh_angular_velocity": 0.1582, "bshin_angular_velocity": -0.0887, "bfoot_angular_velocity": -0.0038, "fthigh_angular_velocity": 0.0233, "fshin_angular_velocity": 0.0696, "ffoot_angular_velocity": 0.1202}, "action": [-0.113, 0.137, 0.816, -0.492, 0.178, -0.282], "next": {"rootz": -0.0722, "rooty": 0.0462, "bthigh": -0.087, "bshin": 0.0393, "bfoot": 0.3176, "fthigh": -0.0562, "fshin": 0.0678, "ffoot": -0.0738, "rootx_velocity": -0.4666, "rootz_velocity": 0.2966, "rooty_angular_velocity": -0.7715, "bthigh_angular_velocity": -0.494, "bshin_angular_velocity": -0.2951, "bfoot_angular_velocity": 7.4441, "fthigh_angular_velocity": -4.5356, "fshin_angular_velocity": 4.1205, "ffoot_angular_velocity": -1.8826}, "reward": -0.4077}
...
[39 more transitions, one JSON object per line]
\end{lstlisting}
\end{mybox}

\subsection{Example synthesized programs}
\label{app:programs}

One champion from each of the three settings, exactly as recorded.

\begin{itemize}
\item Box~\ref{box:champ-pong}: the Pong controller, scoring $+18$.
\item Box~\ref{box:champ-flappy}: the Flappy Bird controller, reaching the episode cap.
\item Box~\ref{box:champ-halfcheetah}: the HalfCheetah feature map.
\end{itemize}

\begin{mybox}[boxrule=0pt, parbox=false, opacityframe=1, colframe=black, label=box:champ-pong]{Pong controller --- median of three runs --- score $+18$}
\begin{lstlisting}[basicstyle=\ttfamily\scriptsize, breaklines=true, columns=fullflexible, keepspaces=true]
# controller.py
import math

def policy(state):
    objs = objects(state)
    player = objs.get("player")
    
    if player is None:
        return "noop"
    
    # Check if there's a ball to track
    ball = objs.get("ball")
    
    if ball is None:
        # No ball yet; hold center position
        return "noop"
    
    # Predict where the ball will be in the near future
    # The ball moves with velocity (dx, dy), so in ~10 ticks it will move by (10*dx, 10*dy)
    # Lead the ball prediction by a few ticks to account for reaction latency
    lead_ticks = 5  # Predict 5 ticks ahead
    predicted_ball_y = ball.y + ball.dy * lead_ticks
    
    player_center = player.y
    deadzone = 4.0  # Tighter deadzone for more responsive control
    
    if player_center < predicted_ball_y - deadzone:
        # Player is above where the ball will be; move down
        return "down"
    elif player_center > predicted_ball_y + deadzone:
        # Player is below where the ball will be; move up
        return "up"
    else:
        # Player is near the predicted ball position; hold position
        return "noop"
\end{lstlisting}
\end{mybox}
\begin{mybox}[boxrule=0pt, parbox=false, opacityframe=1, colframe=black, label=box:champ-flappy]{Flappy Bird controller --- score 311 at the episode cap}
\begin{lstlisting}[basicstyle=\ttfamily\scriptsize, breaklines=true, columns=fullflexible, keepspaces=true]
# controller.py
import math

def policy(state):
    objs = objects(state)
    bird = objs.get("bird")
    
    if bird is None:
        return "noop"
    
    pipes = objs.all("pipe")
    bird_y = bird.y
    bird_dy = bird.dy
    bird_x = bird.x
    
    # AGGRESSIVE ceiling avoidance: keep bird well away from top
    # At y < 80, never flap; at y < 100 only in emergency
    if bird_y < 80:
        return "noop"
    
    if bird_y < 100 and bird_dy < 0:
        return "noop"
    
    # Hard ground bounds
    if bird_y > 500:
        return "flap"
    
    if pipes:
        # Filter pipes ahead of the bird
        pipes_ahead = [p for p in pipes if p.x > bird_x - 50]
        
        if pipes_ahead:
            pipes_ahead.sort(key=lambda p: p.x)
            
            # Identify pipe pair: upper and lower pipe at similar x
            next_upper = None
            next_lower = None
            
            for i in range(len(pipes_ahead)):
                for j in range(i + 1, len(pipes_ahead)):
                    p1, p2 = pipes_ahead[i], pipes_ahead[j]
                    if abs(p1.x - p2.x) < 30:
                        if p1.y < p2.y:
                            next_upper, next_lower = p1, p2
                        else:
                            next_upper, next_lower = p2, p1
                        break
                if next_upper:
                    break
            
            if next_upper and next_lower:
                # Gap geometry
                upper_bottom = next_upper.y + next_upper.h / 2.0
                lower_top = next_lower.y - next_lower.h / 2.0
                gap_center = (upper_bottom + lower_top) / 2.0
                gap_height = lower_top - upper_bottom
                safe_margin = max(20, gap_height * 0.25)
                
                dist_to_pipe = next_upper.x - bird_x
                
                # Close range: precise emergency control only
                if dist_to_pipe < 80:
                    if bird_y < upper_bottom + safe_margin:
                        return "noop"
                    if bird_y > lower_top - safe_margin:
                        return "flap"
                    return "noop"
                
                # Moderate range: steer toward center, but conservatively
                elif dist_to_pipe < 180:
                    target = gap_center
                    if bird_y > target + 30:
                        return "flap"
                    if bird_dy > 3.0 and bird_y > target - 20:
                        return "noop"
                    return "noop"
                
                # Far range: position in middle of gap proactively
                else:
                    target_y = gap_center
                    if bird_y > target_y + 50:
                        return "flap"
                    if bird_dy > 0.5:
                        return "noop"
                    return "noop"
    
    # No pipe ahead; maintain mid-altitude conservatively
    if bird_y > 280:
        return "flap"
    if bird_dy > 1.5:
        return "noop"
    return "noop"
\end{lstlisting}
\end{mybox}

In the continuous setting the model writes only the feature map; the weights are fitted by
search and never shown to it. The map below returns 14 terms, and the fitted $W$ has one row per
actuator.

\begin{mybox}[boxrule=0pt, parbox=false, opacityframe=1, colframe=black, label=box:champ-halfcheetah]{HalfCheetah feature map $\phi$ --- 14 terms}
\begin{lstlisting}[basicstyle=\ttfamily\scriptsize, breaklines=true, columns=fullflexible, keepspaces=true]
import math

def terms(state, t):
    def g(k):
        try:
            v = state[k][0]
        except Exception:
            return 0.0
        if v is None:
            return 0.0
        try:
            f = float(v)
        except Exception:
            return 0.0
        if not math.isfinite(f):
            return 0.0
        return f

    # raw positions
    rootz = g('rootz')
    rooty = g('rooty')
    bthigh = g('bthigh')
    bshin = g('bshin')
    bfoot = g('bfoot')
    fthigh = g('fthigh')
    fshin = g('fshin')
    ffoot = g('ffoot')

    # velocities
    vx = g('rootx_velocity')
    vz = g('rootz_velocity')
    wy = g('rooty_angular_velocity')
    wbt = g('bthigh_angular_velocity')
    wbs = g('bshin_angular_velocity')
    wbf = g('bfoot_angular_velocity')
    wft = g('fthigh_angular_velocity')
    wfs = g('fshin_angular_velocity')
    wff = g('ffoot_angular_velocity')

    def sat(x, s=1.0):
        return math.tanh(x / s)

    out = []
    # 0: constant bias
    out.append(1.0)
    # 1: root height (posture)
    out.append(sat(rootz, 0.2))
    # 2: root pitch angle
    out.append(sat(rooty, 0.3))
    # 3: sum of back-leg joint angles
    out.append(sat(bthigh + bshin + bfoot, 1.0))
    # 4: sum of front-leg joint angles
    out.append(sat(fthigh + fshin + ffoot, 1.0))
    # 5: forward velocity (drives reward)
    out.append(sat(vx, 1.0))
    # 6: vertical velocity
    out.append(sat(vz, 1.0))
    # 7: pitch angular velocity
    out.append(sat(wy, 2.0))
    # 8: combined back-leg angular velocities
    out.append(sat(wbt + wbs + wbf, 10.0))
    # 9: combined front-leg angular velocities
    out.append(sat(wft + wfs + wff, 10.0))
    # 10: coupling term (posture x forward velocity)
    out.append(sat(rooty * vx, 0.5))
    # 11: periodic gait clock
    out.append(math.sin(0.3 * t))
    # 12: periodic gait clock, second frequency / phase
    out.append(math.cos(0.15 * t))
    # 13: front-back leg antisymmetry (differential drive)
    out.append(sat((fthigh - bthigh) + 0.5 * (fshin - bshin), 1.0))

    return [x if math.isfinite(x) else 0.0 for x in out]
\end{lstlisting}
\end{mybox}

\end{document}